\documentclass{article}

\usepackage[numbers]{natbib}
\usepackage[preprint]{neurips_2022}

\usepackage[dvipsnames]{xcolor}         
\definecolor{linkColor}{rgb}{0.18,0.39,0.62}
\usepackage[utf8]{inputenc} 
\usepackage[T1]{fontenc}    
\usepackage[colorlinks=true,linkcolor=linkColor,citecolor=linkColor,filecolor=linkColor,urlcolor=linkColor]{hyperref}       
\usepackage{url}            
\usepackage{booktabs}       
\usepackage{amsfonts}       
\usepackage{nicefrac}       
\usepackage{microtype}      
\usepackage{tcolorbox}

\usepackage{graphicx}
\usepackage{arydshln}
\usepackage{booktabs}
\usepackage{multirow}
\usepackage{caption}
\usepackage{subcaption}
\usepackage{makecell}
\usepackage{csquotes}
\usepackage{epigraph}

\RequirePackage{algorithm}
\RequirePackage{algorithmic}

\usepackage{multirow}
\usepackage{amsmath}
\usepackage{capt-of}
\usepackage{tabularx}
\usepackage{epsfig}
\usepackage{amssymb}
\usepackage{amsfonts}
\usepackage{booktabs}
\usepackage{scalerel}
\usepackage[inline]{enumitem}
\usepackage{listings}
\usepackage{varwidth}
\usepackage[export]{adjustbox}
\usepackage{tikz}
\usetikzlibrary{tikzmark}

\usepackage{stmaryrd}
\usepackage{bbm}
\usepackage{wrapfig}
\usepackage{pifont}
\usepackage[noabbrev]{cleveref}

\newcommand{\tabincell}[2]{\begin{tabular}{@{}#1@{}}#2\end{tabular}}

\definecolor{deepblue}{rgb}{0,0,0.5}
\definecolor{officeblue}{RGB}{0,102,204}
\definecolor{deepred}{rgb}{0.6,0,0}
\definecolor{deepgreen}{rgb}{0,0.5,0}
\definecolor{mybrickred}{RGB}{182,50,28}

\definecolor{fillcolor}{RGB}{216,217,252}

\usepackage{etoolbox}
\usepackage{framed}

\newif\ifxetexorluatex
\ifxetex
  \xetexorluatextrue
\else
  \ifluatex
    \xetexorluatextrue
  \else
    \xetexorluatexfalse
  \fi
\fi
\newcommand*\quotesize{60} 
\newcommand*{\openquote}
   {\tikz[remember picture,overlay,xshift=-4ex,yshift=-2.5ex]
   \node (OQ) {\fontsize{\quotesize}{\quotesize}\selectfont``};\kern0pt}

\newcommand*{\closequote}[1]
  {\tikz[remember picture,overlay,xshift=4ex,yshift={#1}]
   \node (CQ) {\fontsize{\quotesize}{\quotesize}\selectfont''};}

\colorlet{shadecolor}{white}

\newcommand*\shadedauthorformat{\emph} 

\newcommand*\authoralign[1]{%
  \if#1l
    \def\authorfill{}\def\quotefill{\hfill}
  \else
    \if#1r
      \def\authorfill{\hfill}\def\quotefill{}
    \else
      \if#1c
        \gdef\authorfill{\hfill}\def\quotefill{\hfill}
      \else\typeout{Invalid option}
      \fi
    \fi
  \fi}
{\authoralign{#1}
\ifblank{#2}
   {\def\shadequoteauthor{}\def\yshift{-2ex}\def\quotefill{\hfill}}
   {\def\shadequoteauthor{\par\authorfill\shadedauthorformat{#2}}\def\yshift{2ex}}
\begin{snugshade}\begin{quote}\openquote}
{\shadequoteauthor\quotefill\closequote{\yshift}\end{quote}\end{snugshade}}

\usepackage{amsmath,amsfonts,bm}

\def\eqref#1{equation~\ref{#1}}

\def\1{\bm{1}}

\DeclareMathAlphabet{\mathsfit}{\encodingdefault}{\sfdefault}{m}{sl}
\SetMathAlphabet{\mathsfit}{bold}{\encodingdefault}{\sfdefault}{bx}{n}

\newcommand\kosmos{\textsc{Kosmos-1}}
\newcommand\our{\textsc{Kosmos-2}}

\usepackage{pifont}
\newcommand{\cmark}{{\color{blue}\ding{51}}}%
\newcommand{\xmark}{{\color{black}\ding{55}}}%

\newcommand\ie{\textit{i.e.}}
\newcommand\eg{\textit{e.g.}}

\definecolor{bluecode}{RGB}{0, 150, 199}

\title{\our{}: Grounding Multimodal Large Language Models to the World}

\author{
\vspace{-0.25in} \\
Zhiliang Peng\thanks{~Equal contribution. $\dagger$ Corresponding author.},~~Wenhui Wang\footnotemark[1],~~Li Dong\footnotemark[1],~~Yaru Hao,~~Shaohan Huang,~~Shuming Ma,~~{Furu Wei}$^\dagger$ \\
Microsoft Research \\
{\href{https://aka.ms/GeneralAI}{https://aka.ms/GeneralAI}}
\vspace{-0.4cm}
\\}

\date{}

\begin{document}
\maketitle

\vspace{-0.5em}
\begin{abstract}
We introduce \our{}, a Multimodal Large Language Model (MLLM), enabling new capabilities of perceiving object descriptions (\eg, bounding boxes) and grounding text to the visual world.
Specifically, we represent refer expressions as links in Markdown, \ie, ``\texttt{[text span](bounding boxes)}'', where object descriptions are sequences of location tokens.
Together with multimodal corpora, we construct large-scale data of grounded image-text pairs (called \textsc{GrIT}) to train the model.
In addition to the existing capabilities of MLLMs (\eg, perceiving general modalities, following instructions, and performing in-context learning), \our{} integrates the grounding capability into downstream applications.
We evaluate \our{} on a wide range of tasks, including (i) multimodal grounding, such as referring expression comprehension, and phrase grounding, (ii) multimodal referring, such as referring expression generation, (iii) perception-language tasks, and (iv) language understanding and generation.
This work lays out the foundation for the development of Embodiment AI and sheds light on the big convergence of language, multimodal perception, action, and world modeling, which is a key step toward artificial general intelligence.
Code and pretrained models are available at \url{https://aka.ms/kosmos-2}.
\end{abstract}

\vspace{-0.1in}
\begin{figure*}[ht]
\centering
\includegraphics[width=0.9\columnwidth]{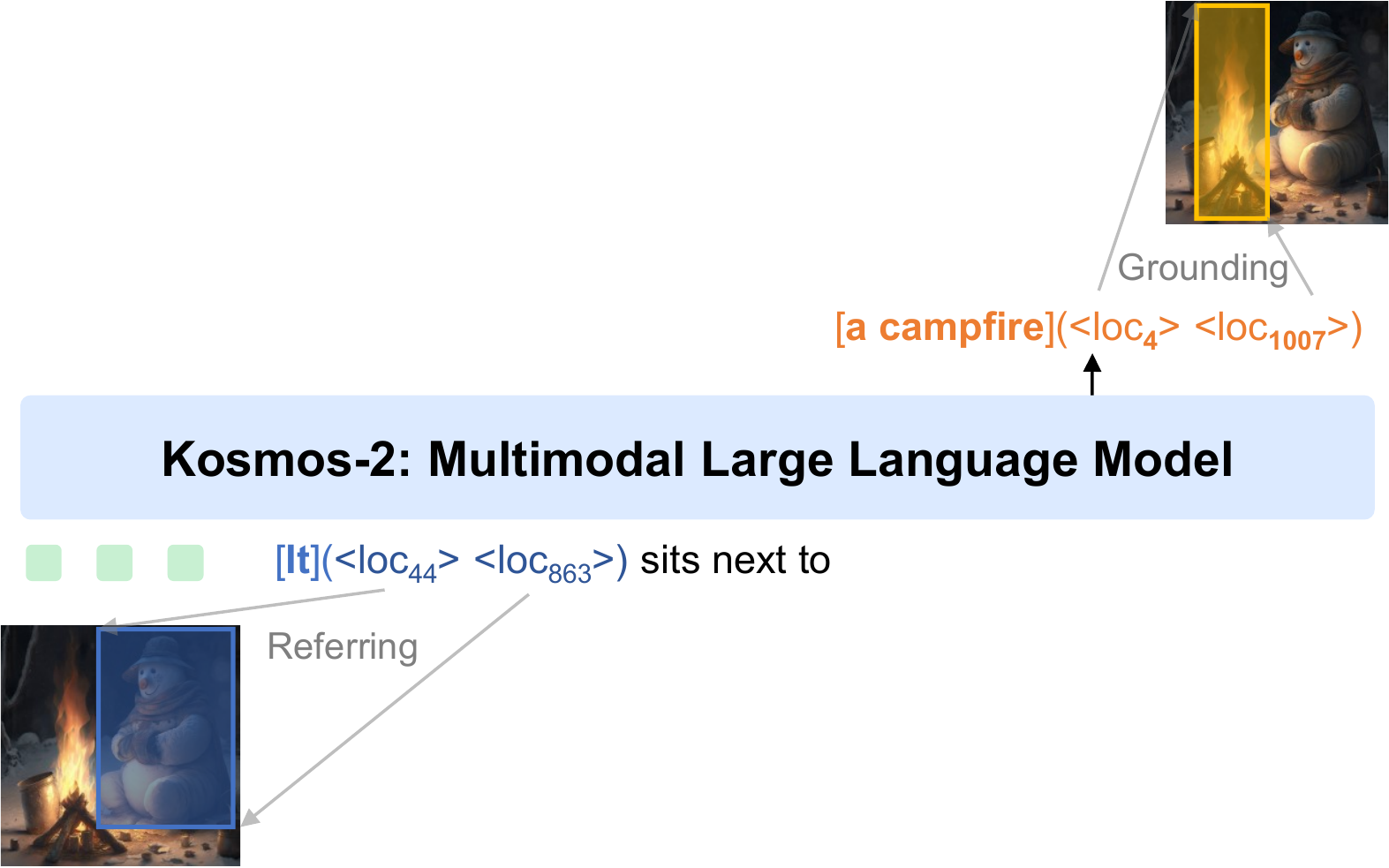}
\caption{
\our{} is a multimodal large language model that has new capabilities of multimodal grounding and referring. \our{} can understand multimodal input, follow instructions, perceive object descriptions (\eg, bounding boxes), and ground language to the visual world.
}
\label{fig:kosmos}
\end{figure*}

\begin{figure*}[hp]
\centering
\includegraphics[width=1.0\textwidth]{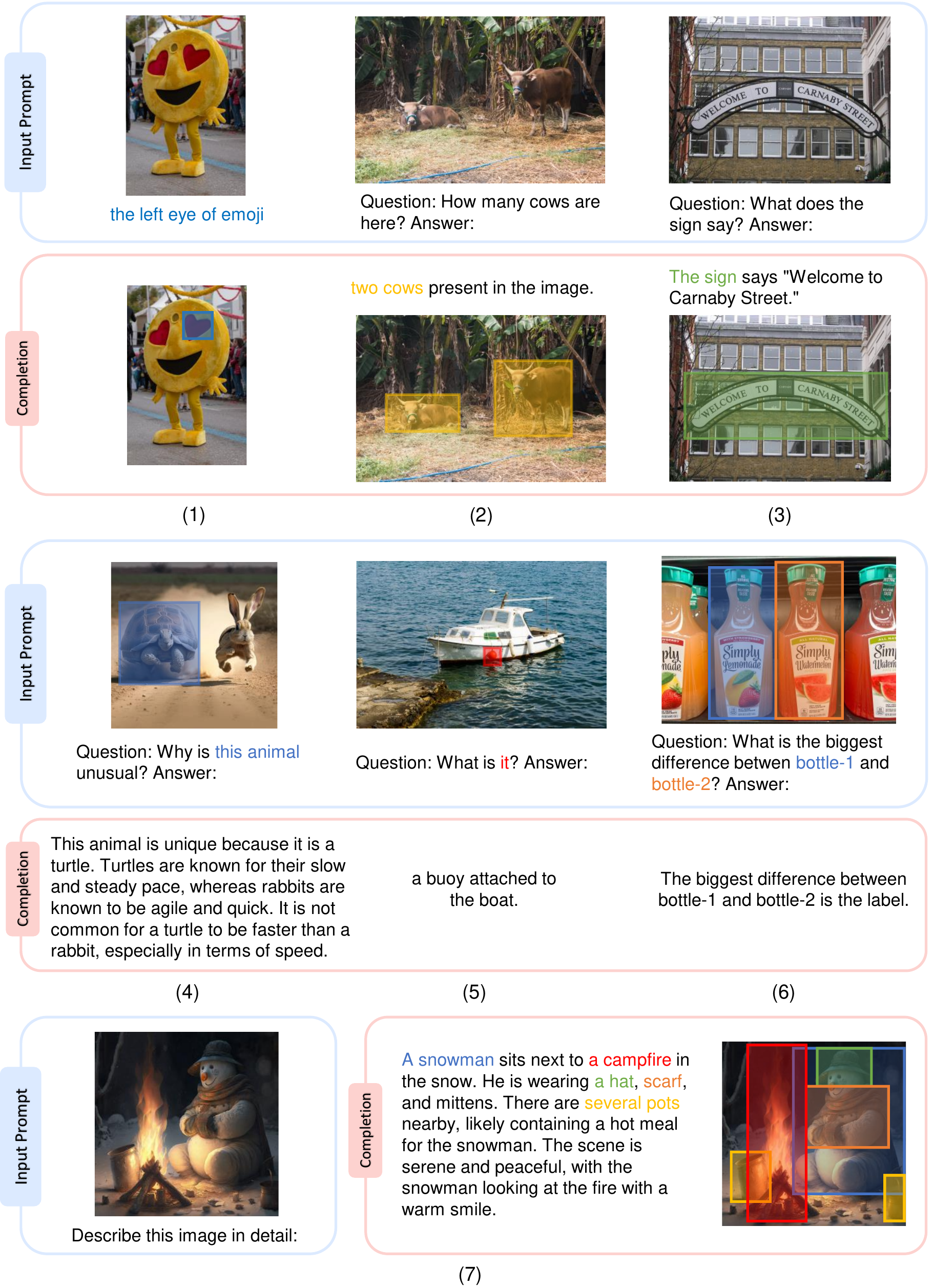}
\caption{
Selected examples generated from \our{}. 
The examples include (1) visual grounding, (2)-(3) grounded question answering, (4)-(6) multimodal referring via bounding boxes, and (7) grounded image captioning.
}
\label{fig:intro:example:1}
\end{figure*}

\newpage

\section{Introduction}
\label{sec:intro}

Multimodal Large Language Models (MLLMs)~\cite{metalm,flamingo,kosmos-1,palm_e,gpt4} have successfully played a role as a general-purpose interface across a wide range of tasks, such as language, vision, and vision-language tasks.
MLLMs can perceive general modalities, including texts, images, and audio, and generate responses using free-form texts under zero-shot and few-shot settings.

In this work, we unlock the grounding capability for multimodal large language models.
Grounding capability can provide a more convenient and efficient human-AI interaction for vision-language tasks.
It enables the user to point to the object or region in the image directly rather than input detailed text descriptions to refer to it, the model can understand that image region with its spatial locations.
Grounding capability also enables the model to respond with visual answers (\ie, bounding boxes), which can support more vision-language tasks such as referring expression comprehension. 
Visual answers are more accurate and resolve the coreference ambiguity compared with text-only responses.
In addition, grounding capability can link noun phrases and referring expressions in the generated free-form text response to the image regions, providing more accurate, informational, and comprehensive answers.

We introduce \our{}, a multimodal large language model with grounding capability built upon \kosmos{}.
\our{} is a Transformer-based causal language model and is trained using the next-word prediction task.
In order to unlock the grounding capability, we construct a web-scale dataset of grounded image-text pairs, and combine it with the multimodal corpora in \kosmos{} to train the model.
The grounded image-text pairs are built upon a subset of image-text pairs from LAION-2B~\cite{laion5b} and COYO-700M~\cite{coyo700m}.
We construct a pipeline to extract and link the text spans (\ie, noun phrases and referring expressions) in the caption to the spatial locations (\eg, bounding boxes) of its corresponding objects or regions in the image.
We convert the spatial coordinates of the bounding boxes to a sequence of location tokens, which is then appended after its respective text spans.
The data format serves as a ``\textit{hyperlink}'' to connect the objects or regions of the image to the caption.

Experimental results demonstrate that \our{} not only achieves competitive performance on language and vision-language tasks evaluated in \kosmos{}, but also achieves impressive performance on grounding tasks (phrase grounding and referring expression comprehension) and referring tasks (referring expression generation).
As shown in Figure~\ref{fig:intro:example:1}, integrating the grounding capability enables \our{} to be used for more downstream tasks, such as grounded image captioning, and grounded visual question answering.

\section{Construction of Web-Scale Grounded Image-Text Pairs (\textsc{GrIT})}

\begin{figure*}[t]
\centering
\includegraphics[width=1.0\textwidth]{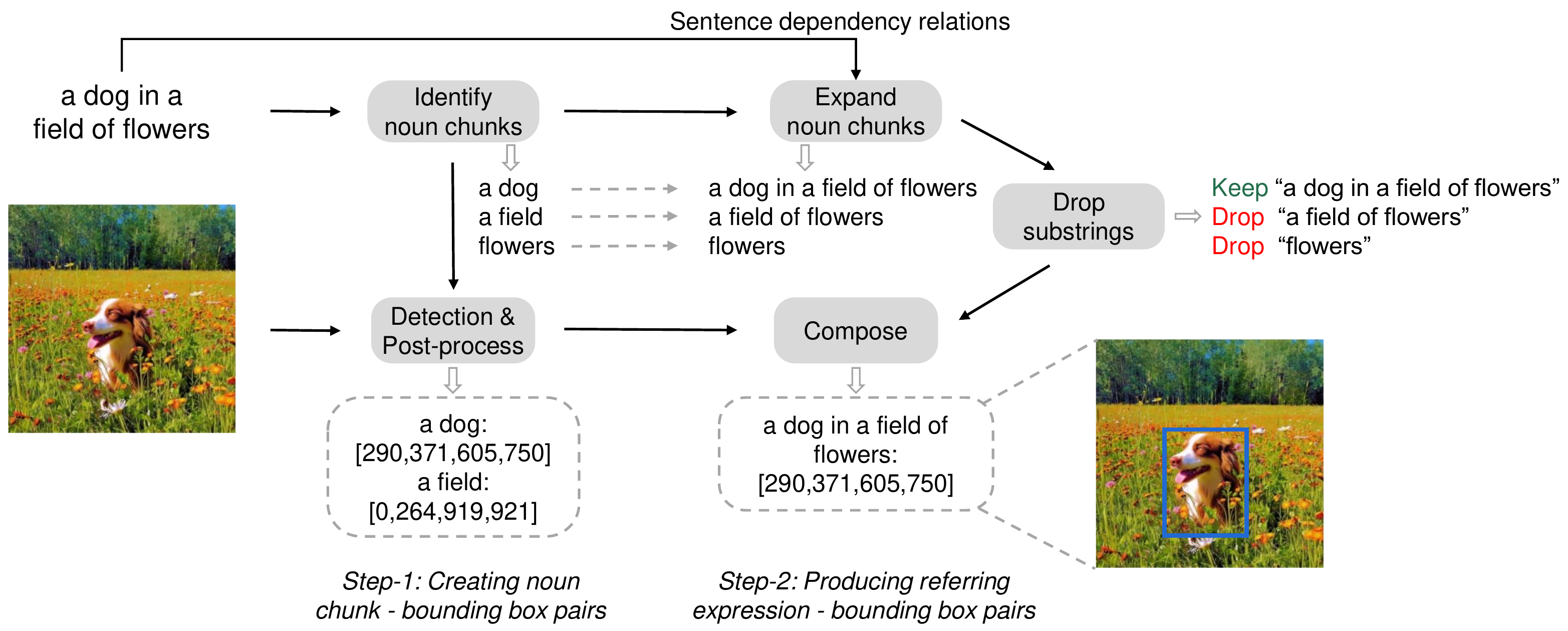}
\caption{
The pipeline of constructing web-scale grounded image-text pairs.
}
\label{fig:data:generate}
\end{figure*}

We introduce \textsc{GrIT}\footnote{A subset of \textsc{GrIT} can be downloaded at \url{https://aka.ms/kosmos-2}.}, a large-scale dataset of \textbf{Gr}ounded \textbf{I}mage-\textbf{T}ext pairs, which is created based on image-text pairs from a subset of COYO-700M~\cite{coyo700m} and LAION-2B~\cite{laion5b}).
We construct a pipeline to extract and link text spans (\ie, noun phrases and referring expressions) in the caption to their corresponding image regions.
The pipeline mainly consists of two steps: generating noun-chunk-bounding-box pairs and producing referring-expression-bounding-box pairs. 
We describe these steps in detail below:

\paragraph{Step-1: Generating noun-chunk-bounding-box pairs}
Given an image-text pair, we first extract noun chunks from the caption and associate them with image regions using a pretrained detector.
As illustrated in Figure~\ref{fig:data:generate}, we use spaCy~\cite{spacy} to parse the caption (``\textit{a dog in a field of flowers}") and extract all noun chunks (``\textit{a dog}'', ``\textit{a field}'' and ``\textit{flowers}'').
We eliminate certain abstract noun phrases that are challenging to recognize in the image, such as ``\textit{time}'', ``\textit{love}'', and ``\textit{freedom}'', to reduce potential noise.
Subsequently, we input the image and noun chunks extracted from the caption into a pretrained grounding model (\eg, GLIP~\cite{glip}) to obtain the associated bounding boxes. 
Non-maximum suppression algorithm is applied to remove bounding boxes that have a high overlap with others, even if they are not for the same noun chunk. 
We keep noun-chunk-bounding-box pairs with predicted confidence scores higher than 0.65.
If no bounding boxes are retained, we discard the corresponding image-caption pair.

\paragraph{Step-2: Producing referring-expression-bounding-box pairs}

In order to endow the model with the ability to ground complex linguistic descriptions, we expand noun chunks to referring expressions.
Specifically, we use spaCy to obtain dependency relations of the sentence.
We then expand a noun chunk into a referring expression by recursively traversing its children in the dependency tree and concatenating children tokens with the noun chunk. 
We do not expand noun chunks with conjuncts.
For noun chunks without children tokens, we keep them for the next process.
In the example shown in Figure~\ref{fig:data:generate}, the noun chunk `\textit{a dog}' can be expanded to ``\textit{a dog in a field of flowers}'', and the noun chunk `\textit{a field}' can be expanded to ``\textit{a field of flowers}''.

Furthermore, we only retain referring expressions or noun chunks that are not contained by others.
As shown in Figure~\ref{fig:data:generate}, we keep the referring expression ``\textit{a dog in a field of flowers}'' and drop ``\textit{a field of flowers}'' (as it is entailed by ``\textit{a dog in a field of flowers}'') and `\textit{flowers}'.
We assign the bounding box of the noun chunk (`\textit{a dog}') to the corresponding generated referring expression (``\textit{a dog in a field of flowers}'').

\begin{table}[t]
\centering
\resizebox{\textwidth}{!}{
\begin{tabular}{lrrrc}
\toprule
\bf Dataset & \bf Images & \bf Objects & \bf Text Spans & \bf Avg Expression Length \\ \midrule
Flickr Entities~\cite{flickr_entity} & 31,783 & 275,775 & 513,644 & - \\
RefCOCOg~\cite{refcocog} & 26,711 & 54,822 & 85,474 & 8.43 \\ 
RefCOCO~\cite{refcoco} & 19,994 & 50,000 & 142,209 & 3.61 \\
RefCOCO+~\cite{refcoco} & 19,992 & 49,856 & 141,564 & 3.53 \\
Visual Genome~\cite{vg} & 108,077 & 4,102,818 & - & - \\
\midrule
\bf \textsc{GrIT} (Ours) &  90,614,680 & 137,349,210 & 114,978,233 & 4.7 \\
\bottomrule
\end{tabular}
}
\vspace{0.2cm}
\caption{Comparison \textsc{GrIT} with existing visual grounding datasets.}
\label{tbl:data:generate:stat}
\end{table}

In the end, we obtain approximately 91M images, 115M text spans, and 137M associated bounding boxes. 
We compare \textsc{GrIT} with existing publicly accessible visual grounding datasets in Table~\ref{tbl:data:generate:stat}.
Data samples of \textsc{GrIT} are shown in the Appendix.

\section{\our{}: A Grounded Multimodal Large Language Model}
\label{sec:methods}

\our{} is a grounded multimodal large language model, which integrates grounding and referring capabilities compared with \kosmos{}.
The model can accept image regions selected by the user using bounding boxes as input, provide visual answers (\ie, bounding boxes), and ground the text output to the visual world.
\our{} adopts the same model architecture and training objective as \kosmos{}.
We add grounded image-text pairs into the training data to endow the model with grounding and referring capabilities.
For a text span (such as noun phrase and referring expression) and its corresponding bounding boxes in a grounded image-text pair, We discretize continuous coordinates of bounding boxes into a sequence of location tokens to encode with text tokens in a unified way.
Then we link the location tokens and their corresponding text span via a ``\textit{hyperlink}'' data format.
The model is trained to establish a mapping between image regions and their corresponding location tokens and connect the image regions with their associated text spans.

\subsection{Grounded Input Representations}

Given a text span and its associated bounding boxes in a grounded image-text pair, we first convert the continuous coordinates of bounding boxes into a sequence of discrete location tokens~\cite{pix2seq}.
For an image with width $W$ and height $H$, we evenly divide both the width and height into $P$ segments each. 
$P \times P$ bins are obtained and each bin consists of ($\nicefrac{W}{P}$) $\times$ ($\nicefrac{H}{P}$) pixels.
For each bin, we use a location token to represent the coordinates within that bin.
We use the coordinates of the center pixel of each bin to determine bounding boxes on the image.
In total, $P \times P$ location tokens are introduced, and these tokens are added to word vocabulary to enable unified modeling with texts.

The bounding box can be represented using its top-left point ($x_1$, $y_1$) and bottom-right point ($x_2$, $y_2$).
We discretize the top-left and bottom-right corner points to location tokens, respectively.
We concatenate the top-left location token \texttt{<loc$_1$>}, the bottom-right location token \texttt{<loc$_2$>}, and special boundary tokens \texttt{<box>} and \texttt{</box>}, to represent a single bounding box: ``\texttt{<box>}\texttt{<loc$_1$><loc$_2$>}\texttt{</box>}''.
If the text span is associated with multiple bounding boxes, we use a special token \texttt{<delim>} to concatenate the location tokens of these bounding boxes: ``\texttt{<box>}\texttt{<loc$_1^i$><loc$_2^i$><delim>...<loc$_1^j$><loc$_2^j$>}\texttt{</box>}''.

Then we arrange the text span and its associated location tokens in a format resembling a ``\textit{hyperlink}'' in markdown.
For the text span with a single bounding box, the resulted sequence is  ``\texttt{<p>} \textit{text span} \texttt{</p>}\texttt{<box>}\texttt{<loc$_1$>}\texttt{<loc$_2$>}\texttt{</box>}'', where \texttt{<p>} and \texttt{</p>} are special tokens indicating the beginning and end of the text span.
The data format tells the model that image regions within the bounding box are associated with the text span.

For the example shown in Figure~\ref{fig:kosmos}, the input representation is:

\begin{minipage}{0.99\columnwidth}
\begin{tcolorbox} 
\small
<s> <image> Image Embedding </image> <grounding> \texttt{<p>} It \texttt{</p>}\texttt{<box>}\texttt{<loc$_{44}$>}\texttt{<loc$_{863}$>}\texttt{</box>} seats next to \texttt{<p>} a campfire \texttt{</p>}\texttt{<box>}\texttt{<loc$_4$>}\texttt{<loc$_{1007}$>}\texttt{</box>}  </s>
\end{tcolorbox}
\vspace{1mm}
\end{minipage}
where \texttt{<s>} and \texttt{</s>} indicate start- and end-of-sequence, and \texttt{<image>} and \texttt{</image>} represent the beginning and end of encoded image embeddings.
\texttt{<grounding>} is a special token to tell the model ground the text output to the visual world.
We map input text tokens and location tokens to embeddings via a lookup table.
Following \kosmos{}, a vision encoder and a resampler module are used to obtain image embeddings for input images.

For language-only data, cross-modal paired data (\ie, image-text pairs), and interleaved multimodal data, we use the same input representations as of \kosmos{}.

\subsection{Grounded Multimodal Large Language Models}

Based on \kosmos{}, \our{} enhances multimodal large language models by incorporating grounding and referring capabilities.
\our{} also uses a Transformer-based causal language model as the backbone and is trained with the next-token prediction task.

In addition to multimodal corpora used in \kosmos{} (including text corpora, image-caption pairs, and interleaved image-text data), we add grounded image-text pairs into training.
The training loss only considers discrete tokens, such as text tokens and location tokens.
The model can learn to locate and understand image regions by their location tokens and the whole image, associate text spans to image regions, and output bounding boxes of the image region using location tokens.

\our{} shows new capabilities of grounding and referring.
The referring capability enables us to point out image regions with bounding boxes.
\our{} can understand the image regions users refer to by the coordinates of bounding boxes.
The referring capability provides a new interaction method.
Different from previous MLLMs~\cite{flamingo,metalm,kosmos-1}, which can only provide text output, \our{} can provide visual answers (\ie, bounding boxes) and ground text output to the image.
The grounding capability enables the model to provide more accurate, informative, and comprehensive responses.
In addition to vision, language, and vision-language tasks evaluated in \kosmos{}, the model can be used for more downstream tasks, such as grounded image-captioning, grounded VQA, referring expression comprehension and generation.

\subsection{Model Training}

\paragraph{Training Setup}
We train the model on newly added grounded image-text pairs, monomodal text corpora, image-caption pairs, and interleaved image-text data. 
Our training process involves a batch size of 419K tokens, consisting of 185K tokens from text corpora, 215K tokens from original and grounded image-caption pairs, and 19K tokens from interleaved data.
We train \our{} for 60k steps, equivalent to around 25 billion tokens. The AdamW optimizer is employed with $\beta=(0.9,0.98)$. 
We set the weight decay to 0.01 and the dropout rate to 0.1. 
The learning rate increases to 2e-4 during the first 375 warm-up steps and linearly decays to zero.
We train the model on 256 V100 GPUs and the training takes approximately one day to complete.
In order to tell the model when to ground text output to the visual world, we prepend the `\texttt{<grounding>}' token to the grounded caption during training.

Following \kosmos{}, the vision encoder has 24 layers with 1,024 hidden size and 4,096 FFN intermediate size.
The multimodal large language model component is a 24-layer \textsc{Magneto} Transformer~\cite{magneto,torchscale} with 2,048 hidden dimensions, 32 attention heads, and 8,192 FFN intermediate size.
The total number of trainable parameters amounts to approximately 1.6B.
The image resolution is set to 224$\times$224 and the patch size is 14$\times$14.
We divide the width and height of the image into 32 bins, with each bin consisting of 7$\times$7 pixels.
A total of 32$\times$32 location tokens are added to the vocabulary.
\our{} uses the weights of \kosmos{} for initialization, the newly added word embeddings of location tokens are initialized randomly.
We update all the parameters during training and instruction tuning.

\paragraph{Instruction Tuning} After the model is trained, we perform instruct tuning to better align \our{} with human instructions.
we combine vision-language instruction dataset (\ie, LLaVA-Instruct~\cite{llava}) and language-only instruction datasets (\ie, Unnatural Instructions~\cite{unnatural} and FLANv2~\cite{flan2}) with the training data to tune the model.
In addition, we construct grounded instruction data by utilizing the pairs of bounding boxes and expressions (\ie, noun phrases, and referring expressions) in \textsc{GrIT}.
Given an expression-bounding-box pair, we use ``<p> \textit{{expression}} </p>'' as the input instruction, and prompt the model to generate the corresponding location tokens of the bounding boxes.
We also use the prompt like ``\texttt{<p>} \textit{It} \texttt{</p>}\texttt{<box>}\texttt{<loc$_1$>}\texttt{<loc$_2$>}\texttt{</box>} \textit{is}'' to ask the model to generate expressions according to its bounding boxes.
Table~\ref{app:corpora:data:refer_template} in Appendix presents more templates.

\section{Evaluation}
\label{sec:eval}

We first evaluate \our{} on multimodal grounding and multimodal referring tasks to assess the new capabilities, and then test the model on language and perception-language tasks evaluated in \kosmos{}.

\begin{itemize}
\item Multimodal grounding
\begin{itemize}
 \item Phrase grounding
 \item Referring expression comprehension
\end{itemize}
\item Multimodal referring
\begin{itemize}
 \item Referring expression generation
\end{itemize}
\item Perception-language tasks
\begin{itemize}
 \item Image captioning
 \item Visual question answering
\end{itemize}
\item Language tasks
\begin{itemize}
 \item Language understanding
 \item Language generation
\end{itemize}
\end{itemize}

\subsection{Multimodal Grounding}
\label{sec:eval:grounding}

In order to evaluate the ability of multimodal grounding, we test \our{} on widely used phrase grounding and referring expression comprehension tasks in a generation manner.
Phrase grounding task requires the model to predict a set of bounding boxes based on one or more given phrases that maybe interrelated within a single caption. 
Referring expression comprehension task encourages the model to locate the object described in a text referring expression within a given image. 

By testing \our{} on these two tasks, we can assess how well the model performs in grounding text descriptions to the visual world, which is crucial for developing advanced AI systems capable of handling complex multimodal tasks.

\begin{figure*}[t]
\centering
\includegraphics[width=0.98\textwidth]{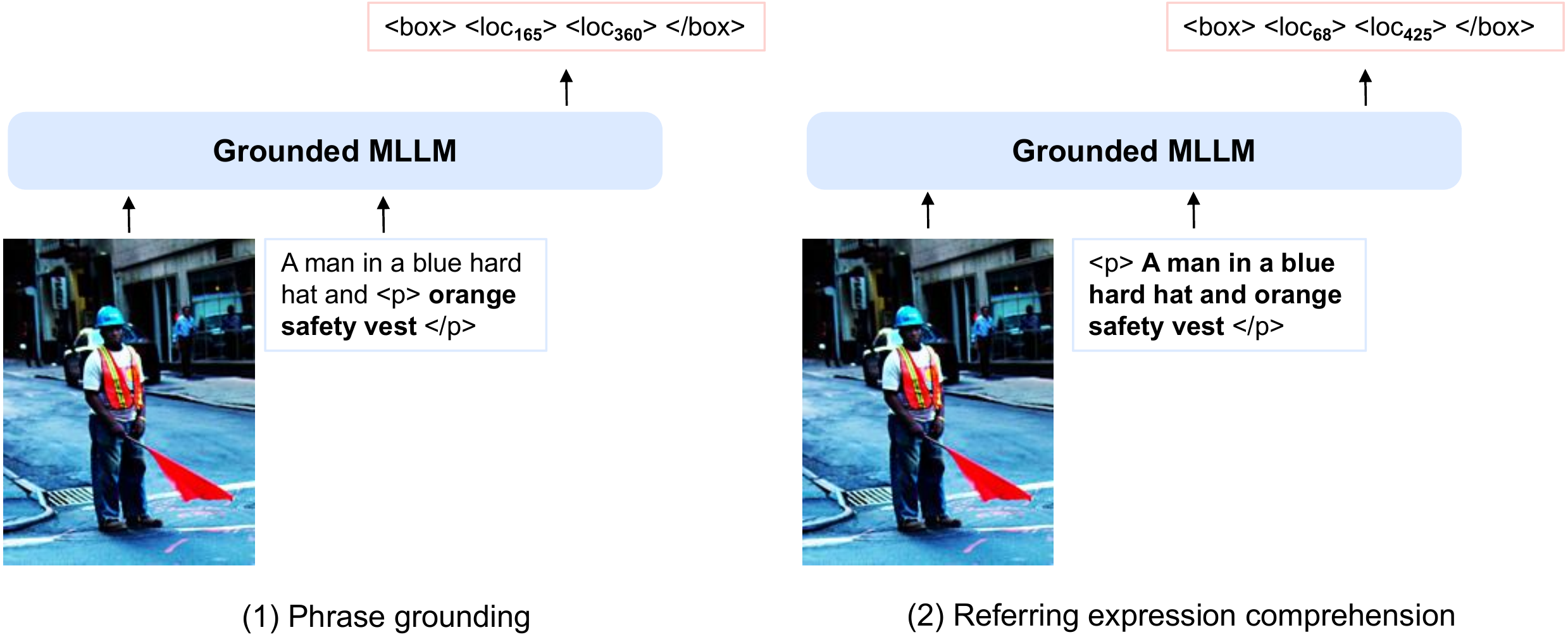}
\caption{
Input format of evaluation on (1) phrase grounding and (2) referring expression comprehension.
}
\label{fig:eval:grd}
\end{figure*}

For both phrase grounding and referring expression comprehension tasks, \our{} is required to generate location tokens which are then converted to bounding boxes for evaluation. 
The input format is ``\texttt{<s>}\texttt{<image>} Image Embedding \texttt{</image>}\texttt{<grounding>}...'', where ``\texttt{<grounding>}'' is used to prompt the model to generate locations tokens.

\subsubsection{Phrase Grounding}

We evaluate phrase grounding task on Flickr30k Entities~\cite{flickr_entity} val and test splits. 
In order to reduce ambiguity, we do not prompt the model with individual phrases; instead, we use the current phrase along with the preceding words as input where preceding words serve as context:
`` ... \texttt{<p>} \{\textit{phrase}\} \texttt{</p>}''.
For the example shown in Figure~\ref{fig:eval:grd}(1), the model needs to predict the locations of phrases ``\textit{A man}'', ``\textit{a blue hard hat}'', ``\textit{orange safety vest}'' and ``\textit{an intersection}'' in the caption ``\textit{A man in a blue hard hat and orange safety vest stands in an intersection.}''. 
To generate the location tokens for the phrase ``\textit{A man}'' that is the beginning of the caption, the prompt is ``\texttt{<p>}\textit{A man}\texttt{</p>}''. 
For the phrase ``\textit{orange safety vest}'', the prompt is ``\textit{A man in a blue hard hat and} \texttt{<p>}\textit{orange safety vest}\texttt{</p>}''.
When multiple men are in the image, the context ``\textit{A man in a blue hard hat and}'' explicitly helps the model locate the object to reduce ambiguity.

We obtain the location tokens in ``\texttt{<box>...</box>}'' from the model response and then covert it into bounding boxes.
The generated bounding box is correct if its intersection over union (IoU) with the ground-truth bounding box is greater than 0.5.
If \our{} generates a location sequence that can not be converted correctly (\eg, ``\texttt{<box><loc$_1$></box>}''), we treat it as a negative sample.
We use \textsc{ANY-BOX} protocol in MDETR~\cite{mdetr}.
We report the R@1, R@5, and R@10 metrics, where R@1/5/10 means calculating the recall using the top 1/5/10 generated bounding boxes. 
If there are fewer than 5 or 10 bounding boxes generated by \our{}, we use all available bounding boxes for the calculation.

\begin{table}[t]
\centering
\begin{tabular}{lcccccccc}
\toprule
\multirow{2}{*}{\textbf{Model}} & \multirow{2}{*}{\textbf{Zero-shot}} & \multicolumn{3}{c|}{\textbf{Val Split}} & \multicolumn{3}{c}{\textbf{Test Split}} \\ \cmidrule(l){3-8} 
 & & R@1 & R@5 & R@10 & R@1 & R@5 & R@10 \\ \midrule
VisualBert~\cite{visualbert} & \xmark & 70.4 & 84.5 & 86.3 & 71.3 & 85.0 & 86.5 \\
MDETR~\cite{mdetr} & \xmark & 83.6 & 93.4 & 95.1 & 84.3 & 93.9 & 95.8\\
GLIP~\cite{glip} & \xmark & 86.7 & 96.4 & 97.9 & 87.1 & 96.9 & 98.1 \\
FIBER~\cite{fiber} & \xmark & 87.1 & 96.1 & 97.4 & 87.4 & 96.4 & 97.6 \\
GRILL~\cite{Jin2023GRILLGV} & \cmark & - & - & - & 18.9 & 53.4 & 70.3 \\ \midrule
\our{} & \cmark & 77.8 & 79.2 & 79.3 & 78.7 & 80.1 & 80.1  \\
\bottomrule
\end{tabular}
\vspace{0.2cm}
\caption{Phrase grounding results on Flickr30k Entities.
We report the R@1, R@5, and R@10 metrics, where R@1/5/10 means calculating the recall using the top 1/5/10 generated bounding boxes. 
}
\vspace{-0.2cm}
\label{tbl:grd:flickr}
\end{table}

\paragraph{Results}
Table~\ref{tbl:grd:flickr} presents results on Flickr30k Entities~\cite{flickr_entity} val and test splits. 
\our{} achieves impressive zero-shot performance and outperforms GRILL~\cite{Jin2023GRILLGV}, which relies on an attached detector, by a large margin. 
Moreover, our model outperforms traditional finetuned  VisualBert~\cite{visualbert} model by 7.4\% R@1 on both val and test splits.
In contrast to other models, \our{} does not involve prior designs (\eg, object queries or proposals), leading to similar results among R@1, R@5, and R@10.
These results demonstrate that \our{} can generate high-quality locations without the need for post-processing redundant locations. 
This capability highlights the effectiveness of our model in handling phrase grounding tasks.

\subsubsection{Referring Expression Comprehension}

We assess the referring expression comprehension task using three well-established datasets: RefCOCO~\cite{refcoco}, RefCOCO+~\cite{refcoco} and RefCOCOg~\cite{refcocog}.
Both RefCOCO and RefCOCO+ were generated through a two-player game, with RefCOCO+ specifically designed to exclude spatial relations, such as ``on the left''. RefCOCOg incorporates spatial relations and features longer expressions on average.
Different from phrase grounding on Flickr30k entities, we measure this task by using referring expression as the input: ``\texttt{<p>} \textit{ referring expression} \texttt{</p>}''. For the example shown in Figure~\ref{fig:eval:grd}(2), the input sequence is ``\texttt{<p>}\textit{A man in a blue hard hat and orange safety vest}\texttt{</p>}''.
Similarly, the predicted bounding box is considered correct only if its IOU with the ground-truth bounding box is greater than 0.5. 
The failed decoded sequence is also treated as a negative sample.
We use the first generated bounding box for the query expression to measure the accuracy.

\begin{table}[t]
\centering
\resizebox{\textwidth}{!}{
\begin{tabular}{lccccccccc}
\toprule
\multirow{2}{*}{\textbf{Model}} & \textbf{Zero-} & \multicolumn{3}{c|}{\textbf{RefCOCO}} & \multicolumn{3}{c|}{\textbf{RefCOCO+}} & \multicolumn{2}{c}{\textbf{RefCOCOg}} \\ \cmidrule(l){3-10} 
 & \textbf{shot} & val & testA & testB & val & testA & testB & val & test \\ \midrule
UNITER~\cite{UNITER} & \xmark & 81.41 & 87.04 & 74.17 & 75.90 & 81.45 & 66.70 & 74.86 & 75.77 \\
MDETR~\cite{mdetr} & \xmark & 87.51 & 90.40 & 82.67 & 81.13 & 85.52 & 72.96 & 83.35 & 83.31 \\
OFA~\cite{ofa} & \xmark & 90.05 & 92.93 & 85.26 & 84.49 & 90.10 & 77.77 & 84.54 & 85.20 \\
FIBER~\cite{fiber} & \xmark & 90.68 & 92.59 & 87.26 & 85.74 & 90.13 & 79.38 & 87.11 & 87.32 \\
VisionLLM~\cite{VisionLLM} & \xmark & 86.7 & - & - & - & - & - & - & - \\ 
GRILL~\cite{Jin2023GRILLGV} & \cmark & - & - & - & - & - & - & - & 47.5 \\ 
\midrule
\our{} & \cmark & 52.32 & 57.42 & 47.26 & 45.48 & 50.73 & 42.24 & 60.57 & 61.65 \\
\bottomrule
\end{tabular}
}
\vspace{0.2cm}
\caption{Referring expression comprehension results on RefCOCO, RefCOCO+ and RefCOCOg.
We report the accuracy metric for all methods.
}
\label{tbl:grd:refcoco}
\end{table}

\paragraph{Results}
Table~\ref{tbl:grd:refcoco} reports referring comprehension results on RefCOCO~\cite{refcoco}, RefCOCO+~\cite{refcoco} and RefCOCOg~\cite{refcocog}.
\our{} also obtains promising zero-shot performance on the comprehension task, significantly outperforming previous zero-shot models on RefCOCOg benchmark.
However, compared to previous finetuned works, \our{} achieves slightly lower performance on RefCOCO and RefCOCO+ than on RefCOCOg.
This discrepancy can be attributed to the data distribution present in RefCOCO and RefCOCO+, where they tend to use a shorter referring expression (\eg, ``left bottom'') during the two-player game.
Hence, one of our future goals is to enhance MLLMs' ability to accurately understand more types of human expressions.

\subsection{Multimodal Referring}

In addition to multimodal grounding tasks, we evaluate the model's ability to understand image regions or objects users refer to via inputting bounding boxes.
Compared with previous multimodal LLMs that can only refer image regions or objects to the model via detailed text descriptions, directly referring to image regions using its bounding boxes is more effective and reduces ambiguity.

We evaluate the model on the referring expression generation task, which aims to generate unambiguous text descriptions of specific objects or regions within the bounding box.
We employ the widely used RefCOCOg dataset~\cite{refcocog} to evaluate the model's performance under both zero-shot and few-shot settings, showcasing its adaptability in different scenarios.

\begin{figure*}[t]
\centering
\includegraphics[width=0.98\textwidth]{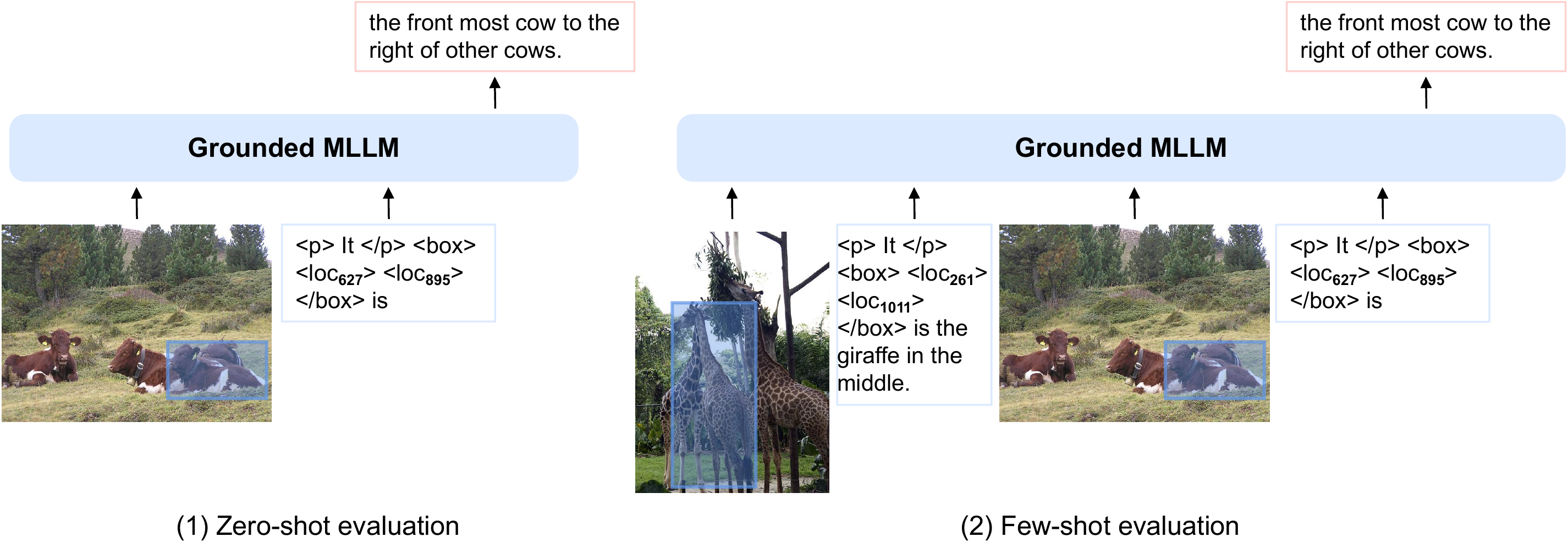}
\caption{
The input format of referring expression generation evaluation under (1) zero-shot and (2) few-shot settings.
The bounding boxes shown in the image are for visualization purposes.
}
\label{fig:eval:gen}
\end{figure*}

\subsubsection{Evaluation Setup}

The model is tasked with generating an associated text description for an object or region given its location tokens of the bounding boxes (\eg, ``\texttt{<box>}\texttt{<loc$_1$>}\texttt{<loc$_2$>}\texttt{</box>}'').
Benefiting from the unified input format, we use ``\texttt{<p>} \textit{It} \texttt{</p>}\texttt{<box>}\texttt{<loc$_1$>}\texttt{<loc$_2$>}\texttt{</box>} \textit{is}'' as prompt to encourage the model to predict its text description.
Figure~\ref{fig:eval:gen} (1) and (2) demonstrate the input format for zero-shot and few-shot referring expression generation, respectively.
Following previous works, we report results using METEOR and CIDEr metrics.
The image resolution is 224$\times$224. Greedy search is used for decoding.

\subsubsection{Results}

Table~\ref{tbl:expgen:refcocog} presents the zero-shot and few-shot results of referring expression generation on RefCOCOg.
We compare \our{} with a finetuned listener-speaker model, which introduces an added reward-based module (SLR).
Our model obtains impressive zero-shot performance on referring expression generation, and even outperforms finetuned SLR by 1.1 CIDEr scores.
Moreover, when prompted with fewshot demonstrations, \our{} shows further improvements, highlighting its in-context learning ability.

\begin{table}[ht]
\centering
\begin{tabular}{llccccc}
\toprule
\multirow{2}{*}{\textbf{Model}} & \multirow{2}{*}{\textbf{Setting}} & \multicolumn{2}{c}{\textbf{RefCOCOg}} \\
 &  & Meteor & CIDEr \\ \midrule
SLR\cite{slr2017} & Finetuning & 15.4 & 59.2 \\
SLR+Rerank\cite{slr2017} & Finetuning & 15.9 & 66.2 \\
\midrule
\multirow{3}{*}{\our{}} & Zero-shot & 12.2 & 60.3 \\
& Few-shot ($k=2$) & 13.8 & 62.2 \\
& Few-shot ($k=4$) & 14.1 & 62.3 \\
\bottomrule
\end{tabular}
\vspace{0.2cm}
\caption{Results of referring expression generation on RefCOCOg.
}
\label{tbl:expgen:refcocog}
\end{table}

\subsection{Perception-Language Tasks}
\label{sec:eval:vl}

In addition to multimodal grounding and referring tasks, we also evaluate \our{} on the vision-language tasks following \kosmos{}.
In particular, we perform zero-shot evaluations on two popular tasks, including image captioning and visual question answering.
Image captioning requires the model to generate a text description of the given image, whereas visual question answering seeks to answer a natural language question based on an image.
In order to have a fair comparison with \kosmos{}, we report results without instruction tuning.

\subsubsection{Evaluation Setup}

For image captioning, we evaluate the model on the widely used Flickr30k \textit{Karpathy split} test set.
We employ beam search for caption generation, with a beam size of 5.
We report results using CIDEr~\cite{cider} metrics evaluated by COCOEvalCap\footnote{\url{https://github.com/salaniz/pycocoevalcap}}.
We use the prompt \textit{``An image of''} to generate the image description.

For visual question-answering, we evaluate zero-shot performance on the test-dev set of VQAv2.
Greedy search is used for decoding.
We report VQA scores obtained from VQAv2 evaluation server\footnote{\url{https://eval.ai/challenge/830/overview}}.
\textit{``Question: \{question\} Answer: \{answer\}''} is used as the prompt for the dataset.
The image resolution is 224$\times$224 for both two tasks.

\subsubsection{Results}

We present the zero-shot performance on Flickr30k and VQAv2 in Table~\ref{tbl:vl:zs-caption-vqa}.
\our{} exhibites comparable overall performance to the \kosmos{}, showing a slight improvement on Flickr30k while experiencing a marginal decrease on VQA.
While \our{} introduces new capabilities of grounding and referring, the model still achieves competitive performance on perception-language tasks.

\begin{table}[ht]
\centering
\begin{tabular}{lccc}
\toprule
\multirow{2}{*}{\textbf{Model}} & \textbf{Flickr30k} & \textbf{VQAv2} \\ \cmidrule(l){2-3} 
 & CIDEr & VQA acc. \\ \midrule
FewVLM~\cite{fewvlm} & 31.0 & - \\
\textsc{MetaLM}~\cite{metalm} & 43.4 & 41.1 \\
Flamingo-3B~\cite{flamingo} & 60.6 & 49.2 \\
Flamingo-9B~\cite{flamingo} & 61.5 & 51.8 \\
\kosmos{}      & 65.2 & 46.7 \\
\our{} & 66.7 & 45.6 \\
\bottomrule
\end{tabular}
\vspace{0.2cm}
\caption{Zero-shot image captioning results on Flickr30k test set and zero-shot visual question answering results on VQAv2 test-dev set.
We report results of \our{} and \kosmos{} without instruction tuning. 
}
\label{tbl:vl:zs-caption-vqa}
\end{table}

\subsection{Language Tasks}
\label{sec:eval:language}

We evaluate \our{} on eight language tasks, such as cloze and completion tasks (StoryCloze, HellaSwag), Winograd-style tasks (Winograd, Winogrande), commonsense reasoning (PIQA), and three SuperGLUE benchmark~\cite{superglue} datasets (BoolQ, CB, and COPA). 
We report the zero-shot results in Table~\ref{tbl:lang:zero_shot}.
Compared with \kosmos{}, \our{} achieves similar performance on StoryCloze, HellaSwag, Winograd, Winogrande, and PIQA, experiences a decrease in performance on CB, but shows improvement on BoolQ and COPA.
In summary, \our{} demonstrates the acquisition of new capabilities while experiencing comparable performance on language tasks. This illustrates the potential of the model in balancing and expanding its skills across different domains.

\begin{table}[ht]
\centering
\resizebox{\textwidth}{!}{
\begin{tabular}{lcccccccc}
\toprule
\textbf{Model} & \textbf{\tabincell{c}{Story \\ Cloze}} & \textbf{\tabincell{c}{Hella \\ Swag}} & \textbf{Winograd} & \textbf{Winogrande} & \textbf{PIQA} & \textbf{BoolQ}  & \textbf{CB} & \textbf{COPA} \\
 \midrule
LLM & 72.9 & 50.4 & 71.6 & 56.7 & 73.2 & 56.4 & 39.3 & 68.0 \\
\kosmos{}  & 72.1 & 50.0 & 69.8 & 54.8 & 72.9 & 56.4 & 44.6 & 63.0 \\
\our{}   & 72.0 & 49.4 & 69.1 & 55.6 & 72.9 & 62.0 & 30.4 & 67.0  \\           
\bottomrule 
\end{tabular}
}
\vspace{0.2cm}
\caption{Zero-shot performance comparisons of language tasks between \our{}, \kosmos{} and LLM. 
LLM uses the same text data and training setup to reimplement a language model as \kosmos{}.
We report results of \our{} and \kosmos{} without instruction tuning.
Results of \kosmos{} and the LLM baseline are from~\cite{kosmos-1}.
}
\label{tbl:lang:zero_shot}
\end{table}

\section{Conclusion}

We present \our{}, a multimodal large language modal, that can ground to the visual world. Specifically, we pre-train \our{} by augmenting the multimodal corpora used in~\kosmos{} with \textsc{GrIT}, a large-scale dataset of Grounded Image-Text pairs, which is created by extracting and associating noun phrases and referring expressions in the caption to the objects or regions in the scene.
\our{} enables new capabilities of perceiving image regions and grounding text output to the visual world, which makes grounding as a foundation capability of MLLMs in many downstream applications.
Experimental results demonstrate that \our{} achieves impressive results on language and vision-language tasks evaluated in \kosmos{}, grounding tasks including phrase grounding and referring expression comprehension, and referring tasks such as referring expression generation.

\section*{Acknowledgement}

Some examples (such as Figure~\ref{fig:kosmos}) are taken from the WHOOPS corpus~\cite{whoops}.

\section*{Ethics Statement}

The model presented in this paper is intended for academic and research purposes.
The utilization of the model to create unsuitable material is strictly forbidden and not endorsed by this work.
The accountability for any improper or unacceptable application of the model rests exclusively with the individuals who generated such content.
We also put Microsoft AI Principles\footnote{\url{https://www.microsoft.com/ai/responsible-ai}} into practice when developing the models.

\bibliographystyle{alpha}
\bibliography{kosmos}

\nocite{cm3}
\nocite{blip2}

\newpage
\appendix

\section{Hyperparameters}
\label{app:hyperparam}

The training hyperparameters of \our{} are listed in Table~\ref{tbl:hyperparam:vl:pt:opt}.

\begin{table}[ht]
\centering
\small
\begin{tabular}{lc}
\toprule
\textbf{Hyperparameters} & \\ \midrule
Image embedding number & 64 \\
Location tokens & 1,024 \\
\midrule
Training steps   &       60,000 \\
Warmup steps     &       375 \\
Optimizer & AdamW \\
Learning rate & 2e-4 \\
Learning rate decay & Linear \\
Adam $\beta$ & (0.9, 0.98) \\
Weight decay & 0.01 \\
\midrule
Batch size of text corpora         &      93  \\
Batch size of original image-caption pairs  & 1,117 \\
Batch size of grounded image-text pairs  & 1,117 \\
Batch size of interleaved data   &    47    \\
\bottomrule
\end{tabular}
\vspace{0.2cm}
\caption{Training hyperparameters of \our{}}
\label{tbl:hyperparam:vl:pt:opt}
\end{table}

The instruction tuning hyperparameters are listed in Table~\ref{tbl:hyperparam:vl:instruct:opt}.
\label{app:hyperparam:inst}

\begin{table}[ht]
\centering
\small
\begin{tabular}{lc}
\toprule
\textbf{Hyperparameters} & \\ \midrule
Training steps                     &       10,000 \\
Warmup steps                      &       375 \\
Learning rate & 1e-5 \\
Batch size of language instruction data     &   117     \\
Batch size of vision-language instruction data  &  351      \\
\tabincell{l}{Batch size of grounded image-text pairs \\ \quad \& grounded instruction data}   &    1404    \\
Batch size of text corpora         &    30    \\
Batch size of interleaved data   &     15     \\

\bottomrule
\end{tabular}
\vspace{0.2cm}
\caption{Instruction tuning hyperparameters of \our{}}
\label{tbl:hyperparam:vl:instruct:opt}
\end{table}

\section{Templates for Grounded Instruction Data}
\label{app:corpora:data:refer_template}

Table~\ref{tbl:corpora:data:refer_template} presents the instruction templates of expression generation based on its associated bounding boxes during instruction tuning.

\begin{table*}[ht]
\centering
\begin{minipage}{0.99\columnwidth}
\begin{tcolorbox} 
\centering
\small
\hspace{-6mm}
\begin{itemize}[leftmargin=7.5mm]
\setlength{\itemsep}{2pt}
\item "What is \texttt{<p>} it \texttt{</p>}\texttt{<box>}\texttt{<loc$_1$>}\texttt{<loc$_2$>}\texttt{</box>}? It is \{\textit{expression}\}."
\item "What is \texttt{<p>} this \texttt{</p>}\texttt{<box>}\texttt{<loc$_1$>}\texttt{<loc$_2$>}\texttt{</box>}? This is \{\textit{expression}\}."
\item "Describe \texttt{<p>} this object \texttt{</p>}\texttt{<box>}\texttt{<loc$_1$>}\texttt{<loc$_2$>}\texttt{</box>}. This object is \{\textit{expression}\}."
\item "\texttt{<p>} It \texttt{</p>}\texttt{<box>}\texttt{<loc$_1$>}\texttt{<loc$_2$>}\texttt{</box>} is \{\textit{expression}\}."
\item "\texttt{<p>} This \texttt{</p>}\texttt{<box>}\texttt{<loc$_1$>}\texttt{<loc$_2$>}\texttt{</box>} is \{\textit{expression}\}."
\item "\texttt{<p>} The object \texttt{</p>}\texttt{<box>}\texttt{<loc$_1$>}\texttt{<loc$_2$>}\texttt{</box>} is \{\textit{expression}\}."
\end{itemize}
\end{tcolorbox}
\vspace{-2mm}
\caption{Instruction templates used for expression generation.}
\label{tbl:corpora:data:refer_template}
\end{minipage}
\end{table*}

\newpage
\section{Examples of \textsc{GrIT}}
\label{app:examples_grounded_pairs}

\begin{figure*}[ht]
\centering
\includegraphics[width=0.5\textwidth]{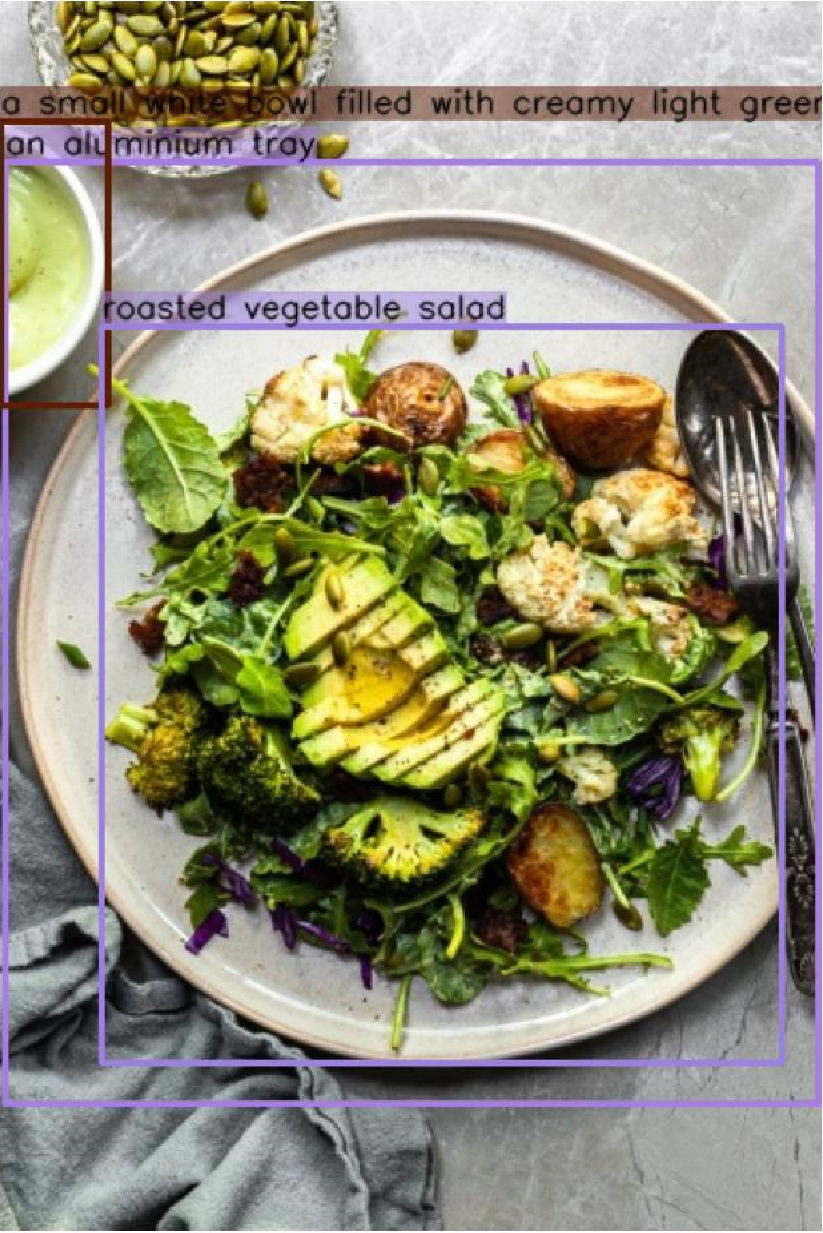}
\caption{Example from \textsc{GrIT}.
Caption: ``\textit{A serving of kale and roasted vegetable salad on an aluminium tray served with a small white bowl filed with creamy light green avocado Caesar dressing}''.
}
\label{fig:data:generate_samples:1}
\end{figure*}

\begin{figure*}[ht]
\centering
\includegraphics[width=0.6\textwidth]{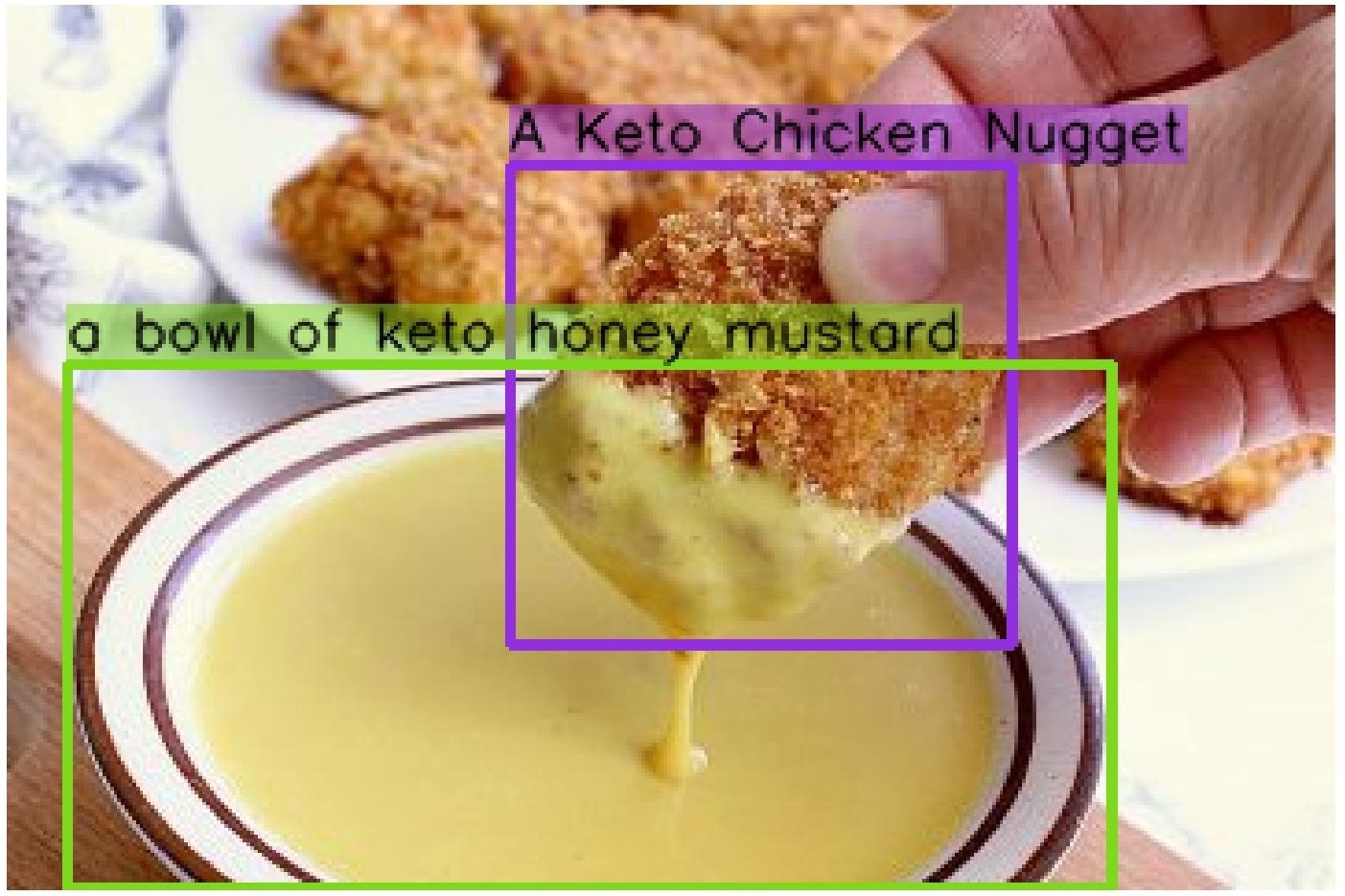}
\caption{Example from \textsc{GrIT}.
Caption: ``\textit{A Keto Chicken Nugget being dipped into a bowl of keto honey mustard.}''.
}
\label{fig:data:generate_samples:2}
\end{figure*}

\begin{figure*}[ht]
\centering
\includegraphics[width=0.85\textwidth]{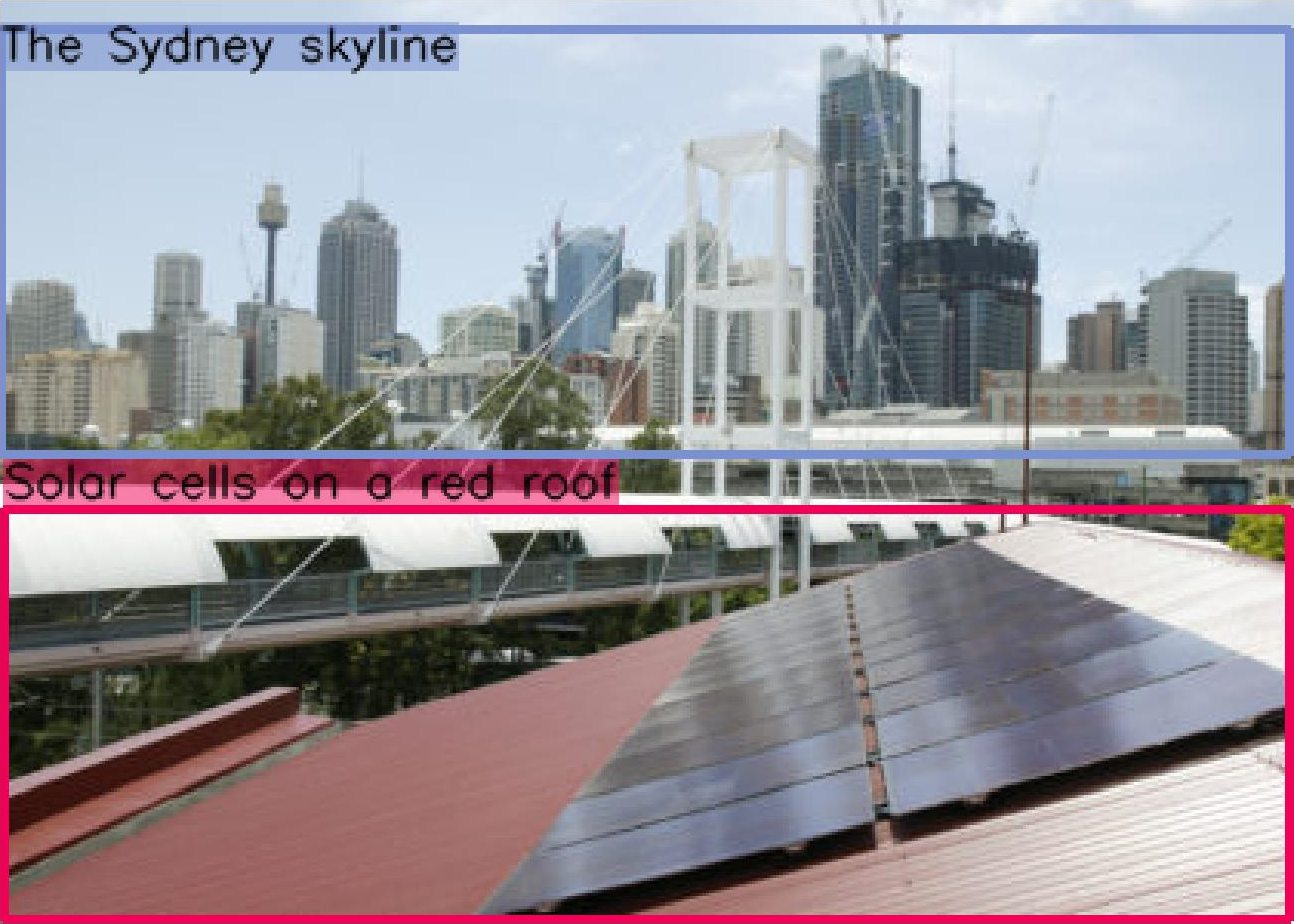}
\caption{Example from \textsc{GrIT}.
Caption: ``\textit{Solar cells on a red roof are in the foreground. The Sydney skyline is in the background.}''.
}
\label{fig:data:generate_samples:3}
\end{figure*}

\begin{figure*}[ht]
\centering
\includegraphics[width=0.9\textwidth]{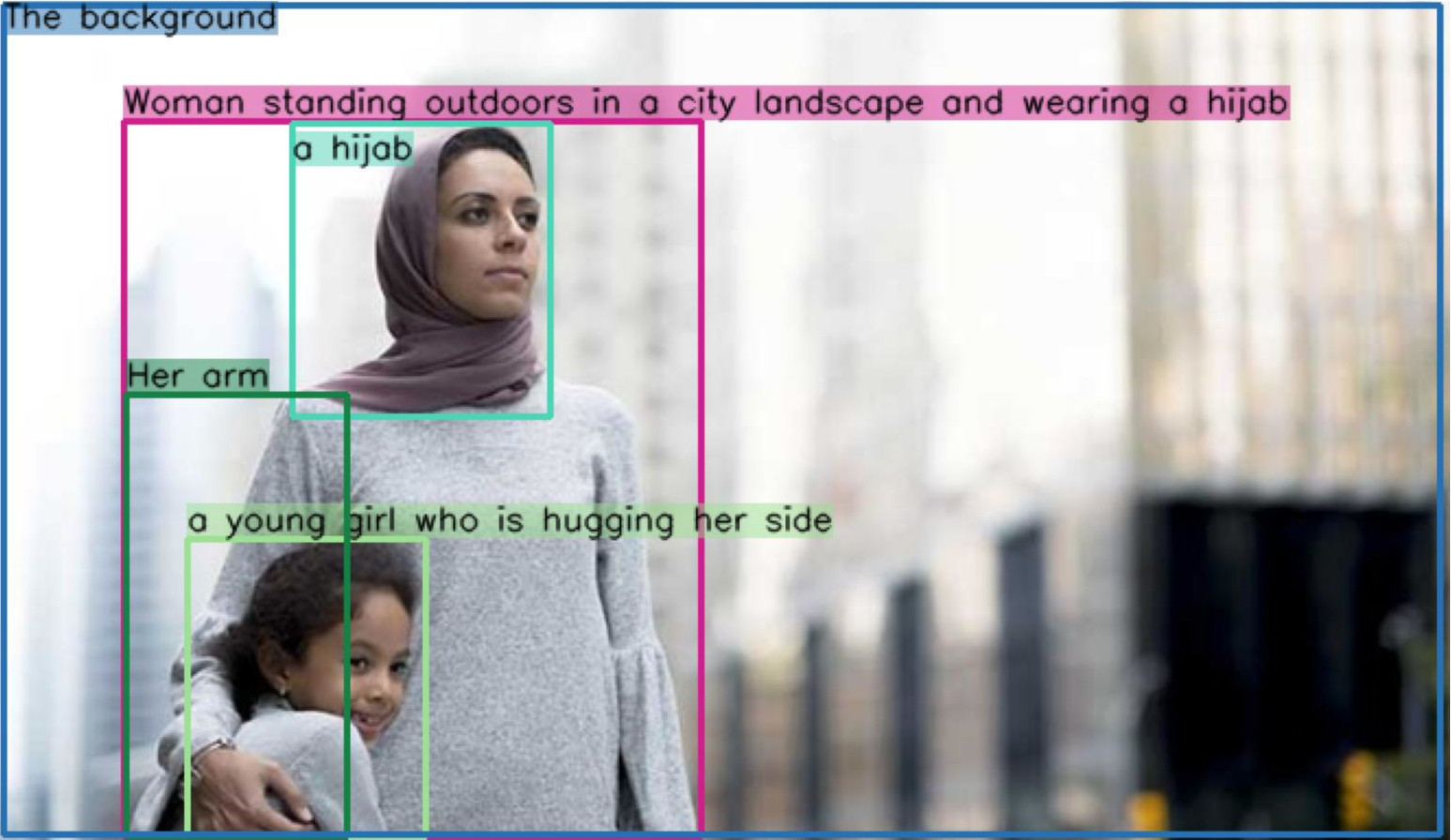}
\caption{Example from \textsc{GrIT}.
Caption: ``\textit{Woman standing outdoors in a city landscape and wearing a hijab. Her arm is around a young girl who is hugging her side. The background is blurred.}''.
}
\label{fig:data:generate_samples:4}
\end{figure*}

We present some examples of the \textsc{GrIT} corpus in
\Cref{fig:data:generate_samples:1,fig:data:generate_samples:2,fig:data:generate_samples:3,fig:data:generate_samples:4}.
The grounded image-text pairs span over various domains and contain different numbers of objects.

\newpage
\section{More Examples of \our{}}

\begin{figure*}[ht]
\centering
\includegraphics[width=0.65\textwidth]{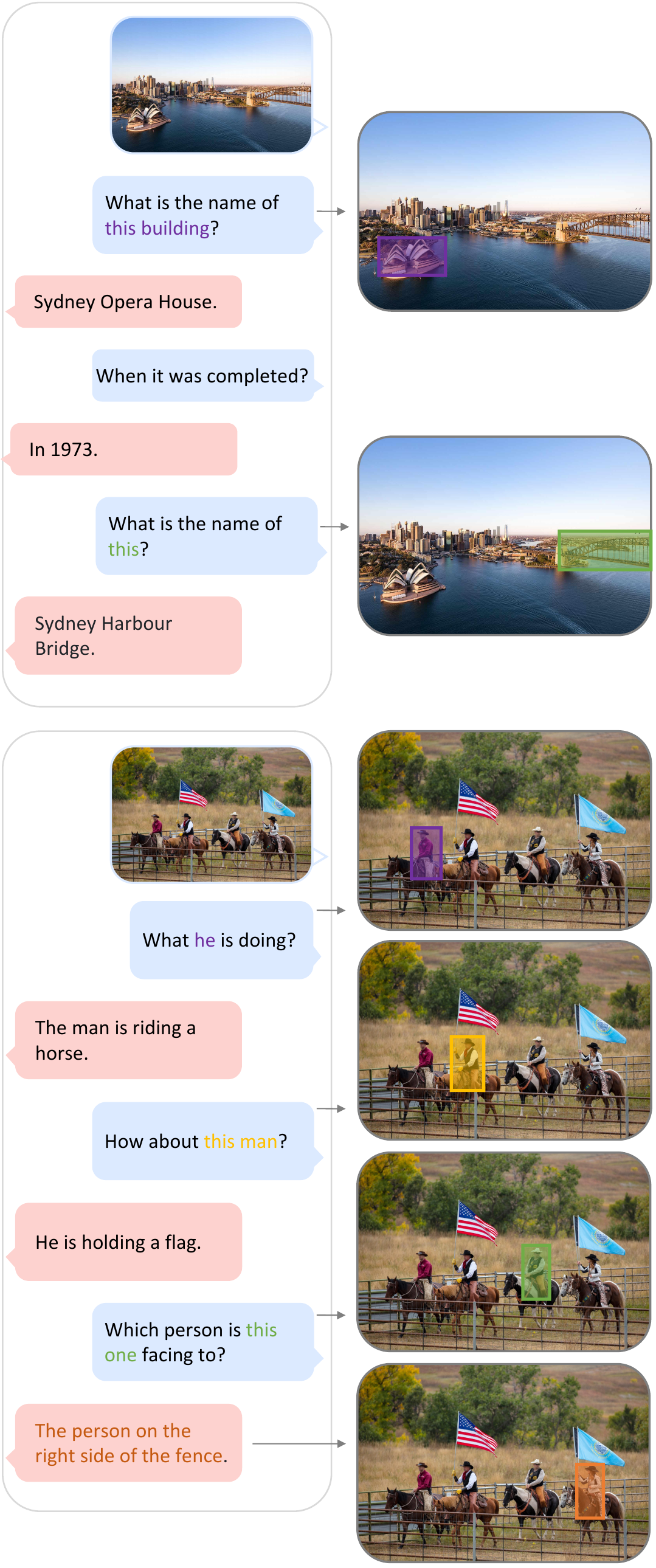}
\caption{
Examples of visual dialogue generated from \our{}.
}
\label{fig:app:example:dialogue}
\end{figure*}

\begin{figure*}[ht]
\centering
\includegraphics[width=0.6\textwidth]{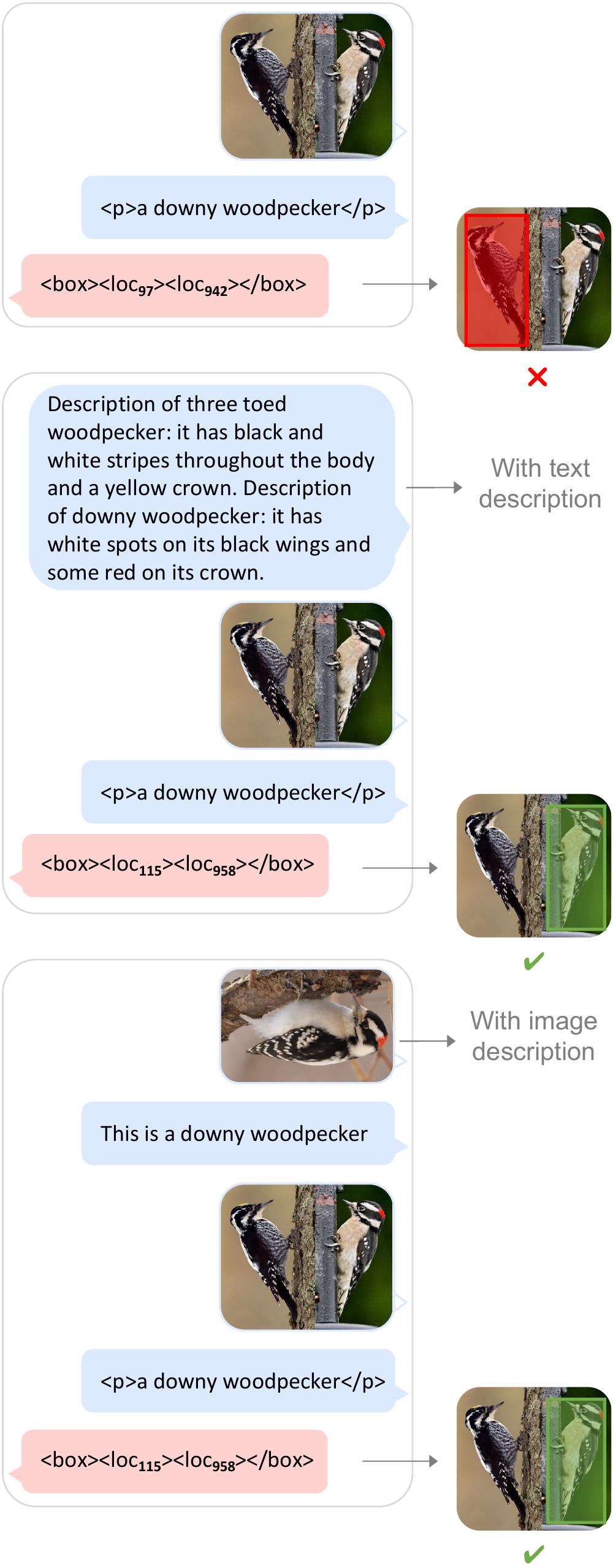}
\caption{
Examples of object detection with multimodal descriptions from \our{}.
}
\label{fig:app:example:descrip}
\end{figure*}

\begin{figure*}[ht]
\centering
\includegraphics[width=1\textwidth]{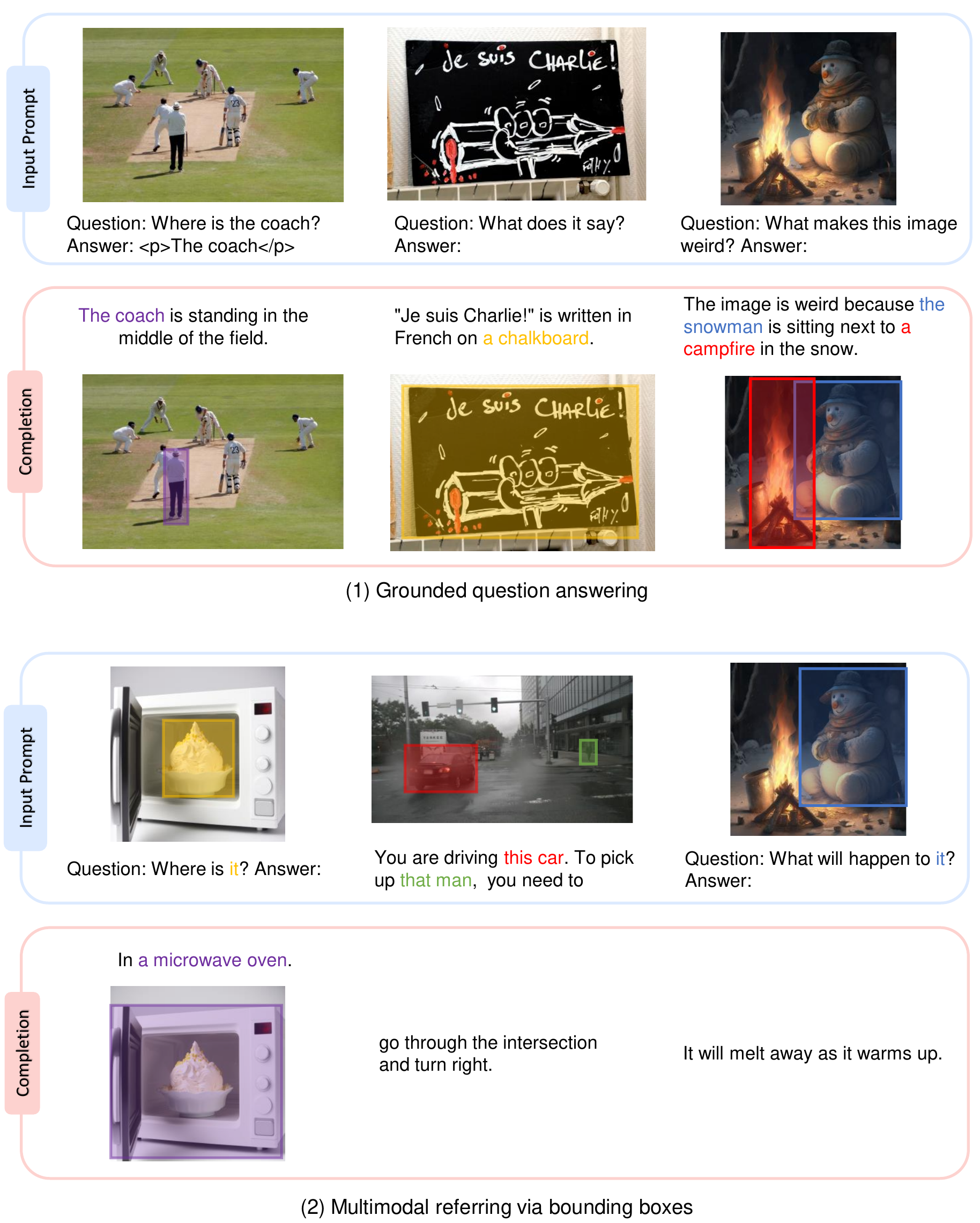}
\caption{
Examples generated from \our{}.
}
\label{fig:app:example:2}
\end{figure*}

\begin{figure*}[hp]
\centering
\includegraphics[width=1\textwidth]{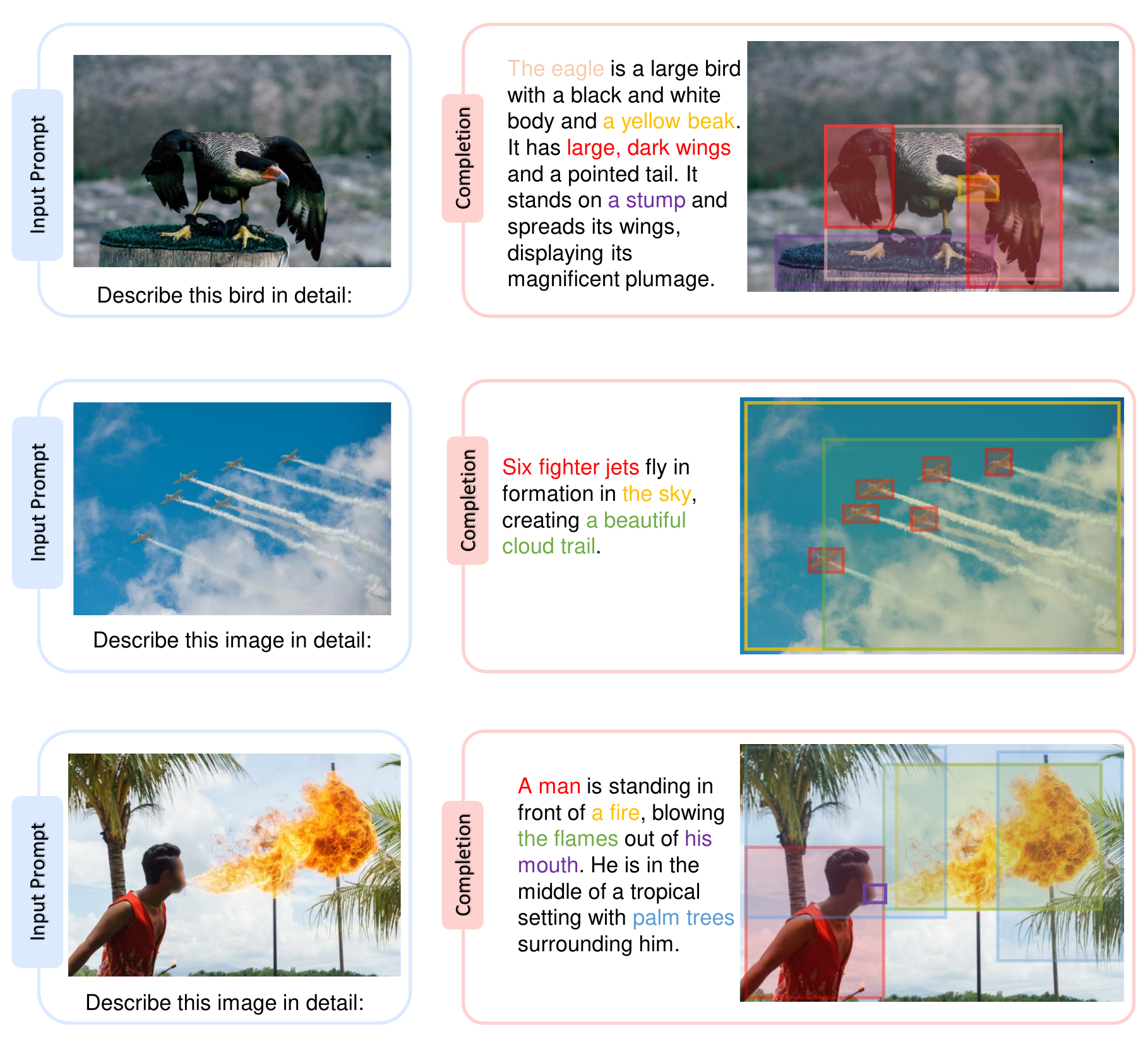}
\caption{
Examples of grounded image captioning generated from \our{}.
}
\label{fig:app:example:3}
\end{figure*}

As illustrated in Figure~\ref{fig:app:example:dialogue}, multimodal referring capability used for visual dialogue can unlock potential in human-AI interaction.
In Figure~\ref{fig:app:example:descrip}, our approach demonstrates its in-context learning ability for fine-grained object detection using both text and image descriptions.
Figure~\ref{fig:app:example:2} and Figure~\ref{fig:app:example:3} showcase more selected examples, including grounded visual question answering, grounded image captioning, and multimodal referring.

\end{document}